\documentclass{opt2022} 

\usepackage[T1]{fontenc}
\usepackage{microtype}
\usepackage{mathtools}
\usepackage{physics}
\usepackage{booktabs}
\usepackage[capitalize,noabbrev]{cleveref}
\usepackage{csquotes}
\usepackage{tikz}
\usepackage{pgffor}
\usepackage{graphics,graphicx}
\usepackage{bm}
\usepackage{siunitx}
\usepackage{multirow,bigdelim}
\usepackage{xkcdcolors}
\usepackage[super]{nth}

\usepackage{todonotes}

\usetikzlibrary{matrix,fit,arrows,arrows.meta,shapes,calc,decorations.pathreplacing,positioning,tikzmark}
\graphicspath{{fig/}}

\newcommand{\Mnist}{MNIST}
\newcommand{\CharFontIm}{CharFontIm}
\NewDocumentCommand{\Cifar}{O{10}}{CIFAR-#1}
\newcommand{\FashionMnist}{Fashion-MNIST}
\newcommand{\Phishing}{Phishing}
\newcommand{\Mushroom}{Mushroom}

\newcommand{\R}{\mathbb{R}}
\newcommand{\N}{\mathbb{N}}

\newcommand*{\elide}{\textup{[\,\dots]}}

\let\Curly\mathcal

\DeclareMathOperator*{\argmin}{arg\,min}

\DeclareMathOperator*{\EX}{\mathbb{E}}
\newcommand{\Exval}[1]{%
  \EX\qty[#1]%
}

\title[Evaluating the Impact of Loss Function Variation]{Evaluating the Impact of Loss Function Variation in Deep Learning for Classification}

\optauthor{%
\Name{Simon Dr{\"a}ger} \Email{simon.draeger@hhu.de}\\
\Name{Jannik Dunkelau} \Email{jannik.dunkelau@hhu.de}\\
\addr Heinrich Heine University D\"usseldorf, Germany}

\begin{document}

\maketitle

\begin{abstract}%
The loss function is arguably among the most important hyperparameters for a neural network.
Many loss functions have been designed to date, making a correct choice nontrivial.
However, elaborate justifications regarding the choice of the loss function are not made in related work.
This is, as we see it, an indication of a dogmatic mindset in the deep learning community which lacks empirical foundation.
In this work, we consider deep neural networks in a supervised classification setting and analyze the impact the choice of loss function has onto the training result.
While certain loss functions perform suboptimally, our work empirically shows that under-represented losses such as the KL Divergence can outperform the State-of-the-Art choices significantly, highlighting the need to include the loss function as a tuned hyperparameter rather than a fixed choice.
\end{abstract}


\section{Introduction}\label{sec:introduction}

In recent years, Deep Neural Networks (DNNs) such as Convolutional Neural Networks (CNNs;~\cite{lecun1989backpropagation}) have been used extensively for learning tasks, including classification.
In a supervised setting, neural networks vitally require an understanding of goodness of their prediction which they receive via the \emph{loss function}.
Many loss functions have been proposed so far, some of which are tailored to specific problems~\cite{wang2019imae,diakogiannis2020resunet,demirkaya2020exploring,feng2018wing,lin2017focal}
while we can more generally
distinguish between loss functions for \emph{classification} and loss functions for \emph{regression}~\cite{Muthukumar2021ClassificationVR}.
In this work, we are only concerned with the classification case.

Usually, researchers select a loss function and assume it throughout their publication.
We highlight the use of seemingly axiomatic statements such as
\enquote{We adopt Mean-Square Error (MSE) as the loss function \elide{}, who [sic] could be rather sensitive to outliers.}~\cite{wang2016studying},
\enquote{The \emph{Cross Entropy} is used as loss function \elide{}}~\cite{tabik2017snapshot} and
\enquote{For parameter optimization, we use the Adam optimizer with cross-entropy loss function.}~\cite{an2020ensemble}.
Some work~\cite{chauhan2018convolutional,ciresan2011flexible} is cited hundred- and thousand-fold but does not mention which loss measure was used.

The loss measure arguably does not receive extensive attention with regards to optimization in deep learning research but is rather subject to a dogmatic choice.
We opine however, that this dogmatic mindset is misplaced in deep learning
due to it being deeply ingrained into the learning process~\cite{yessou2020comparative}
Thus, we conjecture that a poor choice of loss measure leads to converse effects like a bad loss surface, local optima and stalled or slowed optimization.
This can in turn lead to an increase in required epochs and model complexity which pose a higher computational demand on the computing infrastructure.
In this analysis, we vary the loss function across our experiments in order to explore the impact this has on the training outcome.


\section{Considered Loss Functions}%
\label{sec:loss-intro}

Consider the scenario of classifying data points from a data set \(\Curly{D} = \{(x_i, y_i)\}_{i = 1}^N\) into \(c \in \N\) classes.
Using a neural network, a function \(f_\theta\) that maps each data point onto its corresponding class is learned over a set of weights \(\theta\).
To do so, the predicted output and its class are compared by a loss function \(\Curly{L}\) which indicates how good the prediction was.
Subsequently, the loss function is derived wrt.\ \(\theta\) in order to move \(\theta\) into the direction of steepest descent~\cite{rumelhart1985learning}.
Formally, we are interested in
\[\argmin_{\theta \,\in\, \Theta}\ \EX_{x, y}\qty[\Curly{L}(f_\theta(x), y)].\]
However, we do not know the data's joint distribution \(P(\Curly{X}, \Curly{Y})\) and thus have to resort to the expected value's empirical formulation
\[\argmin_{\theta \,\in\, \Theta}\
  \frac{1}{N}\sum\limits_{i = 1}^N \Curly{L}(f_\theta(x_i), y_i).\]
In this work, we consider each label \(y_i\) to be \emph{one-hot encoded}
and thus assume that \(f_\theta(x_i) \in \R^c\) returns a vector of probabilities corresponding to the model's confidence in each class which can be achieved via the Softmax function.

Our analysis incorporates four loss functions (listed in \cref{tab:loss-funs}) that have been widely used in machine learning.
Intuitively, these have different properties, for example, the MAE is Lipschitz continuous with Lipschitz constant \(L = 1\)~\cite{qi2020mean}.
Besides that, it inherits the absolute value function's discontinuity at 0.
Because the MSE squares its arguments component-wise, large differences in the components lead to outliers being penalized stronger.
The MAE, on the other hand, is insensitive to outliers because it weighs all data points equally~\cite{chai2014root}.
The CCE's gradients tend to be steep which is conjectured to be a reason it converges quickly~\cite{bosman2020visualising}.
Both CCE and KLD are not symmetric whereas the MSE and MAE are.
The CCE and KLD have information theoretic interpretations: according to \citet{murphy2012machine},
the CCE is the mean number of bits required 
to encode a source distribution (label distribution) using a model distribution (prediction distribution).
Furthermore, the KLD decomposes into
\begin{equation}
  \Curly{L}_\mathrm{KLD}(\hat{y}, y) = \Curly{L}_\mathrm{CCE}(\hat{y}, y) - \mathbb{H}(y)
\end{equation}
with \(\mathbb{H}(\cdot)\) being the entropy~\cite{murphy2012machine}.
Therefore, the KLD is the mean number of \emph{additional} bits required to encode the data.
\begin{table}[tb]
  \centering
  \caption{Loss functions used in this work. Models using these are implemented, trained and evaluated in the Experiments section.}%
  \label{tab:loss-funs}
  \begin{tabular}{ lll }
    \toprule
    \textbf{Loss Function}      & \textbf{Abbreviation} & \textbf{Expression} \\
    \midrule
    Mean Square Error~\cite{bickel2015mathematical}           & MSE                   & \(\sum\limits_{k = 1}^c \left(\hat{y}^{(k)} - y^{(k)}\right)^2\)                       \\
    Mean Absolute Error~\cite{ghosh2017robust}         & MAE                   & \(\sum\limits_{k = 1}^c \abs{\hat{y}^{(k)} - y^{(k)}}\)                                \\
    (Categorical) Cross-Entropy~\cite{murphy2012machine} & (C)CE                 & \(\sum\limits_{k = 1}^c -y^{(k)}\log\left(\hat{y}^{(k)}\right)\)                       \\
    Kullback-Leibler Divergence~\cite{kullback1951information} & KLD                   & \(\sum\limits_{k = 1}^c y^{(k)}\log\left(\frac{y^{(k)}}{\hat{y}^{(k)}}\right)\)  \\
    \bottomrule
  \end{tabular}
\end{table}

\section{Related Work}\label{sec:related-work}



Similar to our work, \citet{janocha2017loss} compared 12 loss functions in a classification regime.
They investigated synthetic data as well as MNIST~\cite{lecun1989backpropagation}.
The experimental results obtained by \citet{janocha2017loss} and \citet{demirkaya2020exploring} yielded intriguing insight into loss functions for sufficiently complex data sets: less common, non-classical losses such as the \emph{Tanimoto loss}~\cite{diakogiannis2020resunet} and the \emph{Correct-Class Quadratic Loss} (CCQL;~\cite{demirkaya2020exploring}) achieved very good performances.
In an effort to improve the MAE, \citet{wang2019imae} proposed the Improved Mean Absolute Error (IMAE) which, compared to the MAE, led to an increase in training accuracy by 18.8 \% on {\Cifar}.

Despite the success in employing the IMAE, Tanimoto loss and CCQL, we focus on the most popular loss functions which is why less popular (but well-performing) functions are not incorporated in our analysis.
However, we applied our analysis onto more datasets to measure a more generalized impact.

\section{Experiments}\label{sec:experiments}

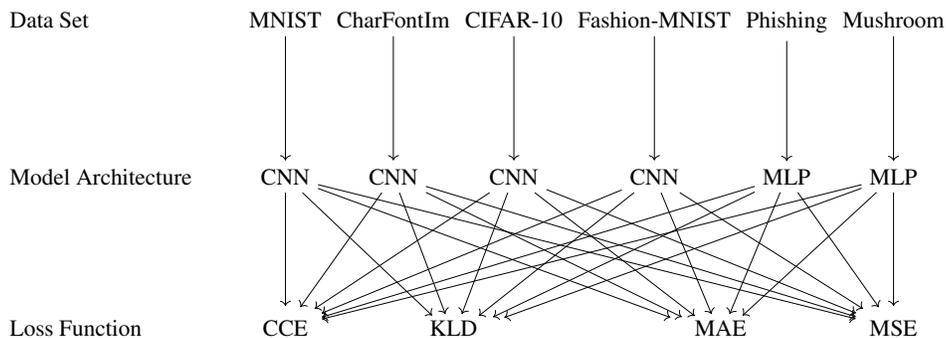
\begin{figure}
  \centering
  \scalebox{0.8}{%
    \begin{tikzpicture}[
	every matrix/.style={column sep=0mm,row sep=2cm},
	node/.style={draw,thick,rounded corners},
	to/.style={->,>=stealth',shorten >=1pt,semithick,font=\sffamily\footnotesize}
]

	\matrix (table) [
		matrix of nodes,
		nodes in empty cells,
		column 1/.style={anchor=base west}
	]{
		Data Set           &[.7cm] MNIST & CharFontIm & CIFAR-10 & Fashion-MNIST & Phishing & Mushroom \\
		Model Architecture &[.7cm] CNN & CNN & CNN & CNN & MLP & MLP \\
		Loss Function      &[.7cm] \node (cce) {CCE}; &&&&& \node (mse) {MSE}; \\
	};

	\node[fit=(table-3-3)(table-3-4)] (kld) {KLD};
	\node[fit=(table-3-5)(table-3-6)] (mae) {MAE};

	\foreach \i in {2,...,7} {%
		\draw[->] (table-1-\i.south) -- (table-2-\i.north);
		\foreach \fun in {cce,kld,mae,mse} {%
			\draw[->,shorten >= 0.1cm] (table-2-\i) -- (\fun);
		}
	}


\end{tikzpicture}%
  }
  \caption{Our experiments as a network diagram.}
  \label{fig:network-diagram}
\end{figure}

In this section, we present an array of experiments regarding the loss functions from \cref{tab:loss-funs}.
In order to avoid results occurring by stochasticity, we generate 6 outcomes (iterations) for each combination of data set and loss function and compute the respective mean metrics.
\Cref{fig:network-diagram} shows a conceptual diagram of our approach.
Training took place
using 2 \(\times\) NVIDIA RTX 8000 + 12 \(\times\) Intel Xeon + 32 GB RAM.
All models are trained for 100 epochs using the Adam optimizer~\cite{kingma2014adam}.

\paragraph{Data Sets.} We used six data sets as presented in \cref{fig:experim-summary}, three of which are popular benchmarking data sets.
The other three ({\Mushroom}, {\Phishing} and {\CharFontIm}) are known to a lesser extent.
{\Mushroom} and {\Phishing} are tabular data sets.
We performed feature analysis via PCA~\cite{abdi2010pca} and concluded that for {\Mushroom}, nearly 100 \% of the data's variance can be explained by 15 features.
Thus, we reduced {\Mushroom} to its 15 most important features,
which are, by index:
0, 1, 3, 6, 7, 8, 10, 11, 13, 14, 16, 17, 18, 19, 20.

\begin{table}[tb]
  \centering
  \caption{Summary of our data sets, including the train/validation/test split.
    The column PCA indicates whether any feature reduction via PCA took place;
    a squiggly arrow in the Features column indicates the reduction result.
    The Train Split column gives the percentages of the resulting
    split into train/validation/test set, respectively.
  }%
  \label{fig:experim-summary}
\begin{tabular}{lrrrlrr}
  \toprule
  {Data Set}                               & \(N\) & \(c\) & Features                   & PCA       & Batch Size & Train Split (\%) \\
  \midrule
  {\Mushroom}~\cite{Dua:2019}              & 8,124   & 2       & 22 \(\rightsquigarrow\) 15 & \checkmark & 16         & 50 / 10 / 40     \\
  {\Phishing}~\cite{Dua:2019}              & 11,055  & 2       & 29                         &           & 16         & 50 / 10 / 40     \\
  \addlinespace
  {\Mnist}~\cite{lecun1989backpropagation} & 70,000  & 10      & \(28\times 28\)            &           & 64         & 76 / 10 / 14     \\
  {\FashionMnist}~\cite{xiao2017fashion}   & 70,000  & 10      & \(28\times 28\)            &           & 64         & 76 / 10 / 14     \\
  {\Cifar}~\cite{krizhevsky2009learning}   & 60,000  & 10      & \(32\times 32\times 3\)    &           & 64         & 74 / 10 / 16     \\
  {\CharFontIm}~\cite{Dua:2019}            & 745,000 & 153     & \(20\times 20\)            &           & 128        & 60 / 10 / 30     \\
  \bottomrule
\end{tabular}
\end{table}

\paragraph{Hyperparameters.} We perform a broad hyperparameter search across the learning rate (\(\lambda\)) and \(\ell_2\) weight decay (\(\eta\)).
The model for each combination is trained for 20 epochs.
Consequently, we choose the parameters that led to the highest validation accuracy.
In total, we sample 30 candidates for \(\lambda\) and \(\eta\) from the log-uniform distribution over \([\num{1e-5}, \num{1e-1}]\).
\begin{table}[tb]
  \centering
  \caption{Hyperparameter configuration for each model and loss function. We round each parameter to the nearest most significant digit. \(\lambda^\ast\) and \(\eta^\ast\) denote the best learning rate and weight decay coefficient, respectively.}%
  \label{tab:hyperparam-config}
  \begin{tabular}{ lcccccccc }
    \toprule
    & \multicolumn{2}{c}{MSE} & \multicolumn{2}{c}{MAE} & \multicolumn{2}{c}{CCE} & \multicolumn{2}{c}{KLD} \\
    \cmidrule(lr){2-3}\cmidrule(lr){4-5}\cmidrule(lr){6-7}\cmidrule(l){8-9}
    Data Set & \(\lambda^\ast\) & \(\eta^\ast\) & \(\lambda^\ast\) & \(\eta^\ast\) & \(\lambda^\ast\) & \(\eta^\ast\) & \(\lambda^\ast\) & \(\eta^\ast\) \\
    \midrule
    {\Mushroom}        & \num{3e-3} & \num{2e-3} & \num{3e-3} & \num{1e-4} & \num{9e-4} & \num{2e-5} & \num{5e-4} & \num{9e-4} \\
    {\Phishing}        & \num{7e-5} & \num{4e-5} & \num{1e-5} & \num{2e-5} & \num{3e-5} & \num{1e-5} & \num{7e-4} & \num{4e-3} \\
    \addlinespace
    {\Mnist}           & \num{2e-3} & \num{1e-5} & \num{2e-4} & \num{3e-5} & \num{1e-5} & \num{9e-5} & \num{5e-4} & \num{4e-3} \\
    {\FashionMnist}    & \num{5e-4} & \num{2e-4} & \num{1e-5} & \num{2e-5} & \num{6e-5} & \num{3e-3} & \num{6e-4} & \num{2e-4} \\
    {\Cifar}           & \num{1e-2} & \num{2e-5} & \num{5e-2} & \num{1e-2} & \num{2e-4} & \num{2e-3} & \num{3e-5} & \num{3e-5} \\
    {\CharFontIm}      & \num{2e-5} & \num{3e-3} & \num{9e-3} & \num{3e-2} & \num{4e-5} & \num{1e-3} & \num{1e-4} & \num{4e-5} \\
    \bottomrule
  \end{tabular}
\end{table}

\paragraph{Architectures.} Two convolutional architectures are obtained from related work~\cite{chauhan2018convolutional,ccalik2018cifar} while the one used on {\CharFontIm} is our own.
We chose these because they performed reasonably well on {\Cifar} and {\Mnist}
(80.17 \% and 99.6 \% training accuracy, respectively).
We use a Multi-Layer Perceptron (MLP) architecture for the tabular data sets,
inspired by previous work on categorical classifiers~\cite{vashisth2020chronic,potghan2018multi,desai2021anatomization,wan2018deep,yulita2018multi,das2018heartbeat}.
For visualizations of the CNNs/MLP, we refer to \cref{appendix:architectures}.

\paragraph{Evaluation Metrics.} We use three metrics in order to assess the trained classifiers' performances.
(I) Test accuracy: we observe the progression of accuracy per elapsed epoch.
The final attained test accuracy is compared across all loss functions.
(II) Matthew's Correlation Coefficient (\(\phi\))~\cite{yule1912methods,matthews1975comparison}.
(III) Area Under Curve (AUC)~\cite{collinson1998bombers}.

\paragraph{Results.} \Cref{tab:mean-test-accuracies-ses} shows the mean test accuracies across \(M = 6\) iterations of an experiment.
In \cref{tab:perf-metrics}, we present the attained mean performances for
\(\phi\) and AUC\@.
Our results show that, performance-wise, the KLD outperforms all other loss functions in 4 out of 6 experiments.
Remarkably, the KLD attains 100 \% test accuracy on all 6 iterations for {\Mushroom}.
While the State-of-the-Art for classification (CCE) performs slightly worse than the MSE, it is more stable.
Unexpectedly, the MAE performs best and most stably on {\Mnist} solely.
However, as we transition to more challenging data sets, test accuracy degrades into pathological cases for the MSE and MAE\@; on {\Cifar}, they were only able to ever predict one class (10 \% accuracy).
During training, the MAE almost always kept an offset in test accuracy to
the other losses
(cf.\ \cref{fig:accuracy-progressions} in \cref{appendix:accuracy-progressions}).
We hypothesize that this is due to its bad gradient properties.
On {\Mnist}, its performance was best which we attribute to the possibility that {\Mnist} may be classifiable by learning its conditional median \(\Exval{\abs{f_\theta(x) - y_i} \mid x = x_i}\), \(i \in \{1, \dots, N\}\).


\begin{table}[tb]
  \centering
  \caption{Mean attained test accuracies and corresponding standard error of mean (SEM) after 100 epochs. The best accuracies \(\pm\) SEMs for each row are \textbf{boldfaced}.
    A higher mean accuracy is better. A lower SE is better except if the attained accuracy is pathological. Here, the SE should be understood as a measure of stability across \(M\) iterations of an experiment. 
    }%
  \label{tab:mean-test-accuracies-ses}
  \resizebox{\columnwidth}{!}{\begin{tabular}{lr@{\hskip 0.5mm}c@{\hskip 0.5mm}rr@{\hskip 0.5mm}c@{\hskip 0.5mm}rr@{\hskip 0.5mm}c@{\hskip 0.5mm}rr@{\hskip 0.5mm}c@{\hskip 0.5mm}r}
    \toprule
    Data Set & \multicolumn{3}{c}{MSE} & \multicolumn{3}{c}{MAE} & \multicolumn{3}{c}{CCE} & \multicolumn{3}{c}{KLD} \\
    \midrule
    {\Mushroom}     & \(99.69 \%\) & \(\pm\) & \(0.26 \%\)          & \(95.83 \%\)          & \(\pm\) & \(0.27 \%\)          & \(99.67 \%\)          & \(\pm\) & \(0.10 \%\)          & \(\mathbf{100.00}\%\) & \(\pm\) & \(\mathbf{0.00}\%\) \\
    {\Phishing}     & \(91.41 \%\) & \(\pm\) & \(\mathbf{0.08}\%\) & \(88.70 \%\)          & \(\pm\) & \(0.32 \%\)          & \(90.29 \%\)          & \(\pm\) & \(0.11 \%\)          & \(\mathbf{92.07}\%\)  & \(\pm\) & \(0.26 \%\) \\
    \addlinespace
    {\Mnist}        & \(98.53 \%\) & \(\pm\) & \(0.17 \%\)          & \(\mathbf{99.30}\%\) & \(\pm\) & \(\mathbf{0.01}\%\) & \(98.99 \%\)          & \(\pm\) & \(0.03 \%\)          & \(98.84 \%\)           & \(\pm\) & \(0.04 \%\) \\
    {\FashionMnist} & \(87.96 \%\) & \(\pm\) & \(\mathbf{0.04}\%\) & \(82.04 \%\)          & \(\pm\) & \(0.09 \%\)          & \(88.60 \%\)          & \(\pm\) & \(0.11 \%\)          & \(\mathbf{90.47}\%\)  & \(\pm\) & \(0.11 \%\) \\
    {\Cifar}        & \(10.00 \%\) & \(\pm\) & \(0.00 \%\)          & \(10.00 \%\)          & \(\pm\) & \(0.00 \%\)          & \(\mathbf{78.13}\%\) & \(\pm\) & \(\mathbf{0.28}\%\) & \(70.05 \%\)           & \(\pm\) & \(0.30 \%\) \\
    {\CharFontIm}   & \(11.30 \%\) & \(\pm\) & \(0.00 \%\)          & \(3.37 \%\)           & \(\pm\) & \(1.52 \%\)          & \(69.21 \%\)          & \(\pm\) & \(0.97 \%\)          & \(\mathbf{76.62}\%\)  & \(\pm\) & \(\mathbf{0.02}\%\) \\
    \bottomrule
  \end{tabular}}
\end{table}
\begin{table}[tb]
  \centering
  \caption{Performance metrics for the data sets. The best entries for each row are \textbf{boldfaced}. Recall that a higher \(\phi\) and a higher AUC is better. For {\Mnist}, no AUC column is boldfaced since all attain equal values. For pathological cases (MSE/MAE for {\Cifar} and {\CharFontIm}), no entry is boldfaced.}%
  \label{tab:perf-metrics}
  \centerline{%
    \begin{tabular}{lrrrrrrrr}
      \toprule
      & \multicolumn{2}{c}{MSE} & \multicolumn{2}{c}{MAE} & \multicolumn{2}{c}{CCE} & \multicolumn{2}{c}{KLD} \\
      \cmidrule(rl){2-3}\cmidrule(lr){4-5}\cmidrule(lr){6-7}\cmidrule(l){8-9}
      Data Set & \multicolumn{1}{c}{\(\phi\)} & \multicolumn{1}{c}{AUC} & \multicolumn{1}{c}{\(\phi\)} & \multicolumn{1}{c}{AUC} & \multicolumn{1}{c}{\(\phi\)} & \multicolumn{1}{c}{AUC} & \multicolumn{1}{c}{\(\phi\)} & \multicolumn{1}{c}{AUC} \\
      \midrule
      {\Mushroom}     & 0.99 & 0.99 & 0.91              & 0.94 & 0.99              & 1.00       & \(\mathbf{1.00}\) & \(\mathbf{1.00}\) \\
      {\Phishing}     & 0.82 & 0.96 & 0.77              & 0.94 & 0.80              & 0.96       & \(\mathbf{0.83}\) & \(\mathbf{0.97}\) \\
      \addlinespace
      {\Mnist}        & 0.98 & 0.99 & \(\mathbf{0.99}\) & 0.99 & 0.98              & 0.99              & 0.98              & 0.99 \\
      {\FashionMnist} & 0.86 & 0.98 & 0.80              & 0.97 & 0.87              & \(\mathbf{0.99}\) & \(\mathbf{0.89}\) & \(\mathbf{0.99}\) \\
      {\Cifar}        & 0.00 & 0.50 & 0.00              & 0.50 & \(\mathbf{0.75}\) & \(\mathbf{0.97}\) & 0.66              & 0.95 \\
      {\CharFontIm}   & 0.00 & 0.50 & 0.00              & 0.50 & 0.67 & 0.98       & \(\mathbf{0.75}\) & \(\mathbf{0.99}\) \\
      \bottomrule
    \end{tabular}%
  }
\end{table}

\section{Future Work}\label{sec:future-work}

Our work leaves improvements for future research.
Firstly, we believe that more data sets need to be incorporated into loss function variation analysis.

Secondly, since we only considered classification, our work leaves open the question of how the impact of loss function variation expresses in regression settings.
How significant and nontrivial is the selection of the loss function in regression?
Besides our restriction to four loss functions, we mentioned novel losses such as the IMAE, Tanimoto loss and CCQL which outperformed classical losses.
Therefore, we recommend more work be dedicated to designing custom loss measures.
A hypothesis regarding the KLD's very good performance may be that the class of \(f\)-divergences provides good loss functions.
Examples of \(f\)-divergences are the KLD, the Bhattacharyya divergence~\cite{bhattacharyya1943measure} and Hellinger distance~\cite{hellinger1909neue}.

Thirdly, an expansion to sequence-based models would be interesting.
To date, several well-performing neural architectures, such as the Transformer~\cite{vaswani2017attention}, have been proposed.

\section{Conclusions}\label{sec:conclusion}

In this work, we explored the impact of varying the loss function in different deep learning architectures on the training process and its performance.
We considered the case of classifying data points into \(c\) classes using probabilistic predictions with one-hot encoded labels.
We have demonstrated empirically that on {\Mnist}, the MAE can achieve a better classifier performance than the CCE or KLD.
Our designation is that if rapid convergence is needed, the MSE can be favored for a low amount of classes.
Our main finding is that the KLD outperforms the CCE on 4 out of 6 data sets.
Along with very good test accuracies, two performance metrics are underpinnings of this result.
Thus, our recommendation is to increase the KLD's usage in binary and multi-class classification regimes.
Due to its very bad performance on color images and overall subpar performance, we advise that using the MAE in a classification setting should be considered carefully.
Finally,
we encourage a paradigm shift that steers away from axiomatically fixating loss
functions but considers them part of hyperparameter optimization.

\paragraph*{Acknowledgements.}

Computational infrastructure and support were provided by the Center for Information and Media Technology at Heinrich Heine University D{\"u}sseldorf.

\bibliography{references}

\begin{thebibliography}{40}
\providecommand{\natexlab}[1]{#1}
\providecommand{\url}[1]{\texttt{#1}}
\expandafter\ifx\csname urlstyle\endcsname\relax
  \providecommand{\doi}[1]{doi: #1}\else
  \providecommand{\doi}{doi: \begingroup \urlstyle{rm}\Url}\fi

\bibitem[Abdi and Williams(2010)]{abdi2010pca}
Hervé Abdi and Lynne~J. Williams.
\newblock Principal component analysis.
\newblock \emph{{WIREs} Computational Statistics}, 2\penalty0 (4):\penalty0
  433--459, 2010.
\newblock \doi{https://doi.org/10.1002/wics.101}.

\bibitem[An et~al.(2020)An, Lee, Park, Yang, and So]{an2020ensemble}
Sanghyeon An, Minjun Lee, Sanglee Park, Heerin Yang, and Jungmin So.
\newblock An ensemble of simple convolutional neural network models for {MNIST}
  digit recognition.
\newblock \emph{arXiv preprint arXiv:2008.10400}, 2020.

\bibitem[Bhattacharyya(1943)]{bhattacharyya1943measure}
Anil Bhattacharyya.
\newblock On a measure of divergence between two statistical populations
  defined by their probability distributions.
\newblock \emph{Bull. Calcutta Math. Soc.}, 35:\penalty0 99--109, 1943.

\bibitem[Bickel and Doksum(2015)]{bickel2015mathematical}
Peter~J. Bickel and Kjell~A. Doksum.
\newblock \emph{Mathematical Statistics: Basic Ideas and Selected Topics,
  Volumes {I--II} Package}.
\newblock Chapman and Hall/CRC, 2015.

\bibitem[Bosman et~al.(2020)Bosman, Engelbrecht, and
  Helbig]{bosman2020visualising}
Anna~Sergeevna Bosman, Andries Engelbrecht, and Mard{\'e} Helbig.
\newblock Visualising basins of attraction for the cross-entropy and the
  squared error neural network loss functions.
\newblock \emph{Neurocomputing}, 400:\penalty0 113--136, 2020.

\bibitem[{\c{C}}alik and Demirci(2018)]{ccalik2018cifar}
Rasim~Caner {\c{C}}alik and M.~Fatih Demirci.
\newblock Cifar-10 image classification with convolutional neural networks for
  embedded systems.
\newblock In \emph{2018 IEEE/ACS 15th International Conference on Computer
  Systems and Applications ({AICCSA})}, pages 1--2. IEEE, 2018.

\bibitem[Chai and Draxler(2014)]{chai2014root}
Tianfeng Chai and Roland~R. Draxler.
\newblock Root mean square error ({RMSE}) or mean absolute error ({MAE})? --
  arguments against avoiding {RMSE} in the literature.
\newblock \emph{Geoscientific Model Development}, 7\penalty0 (3):\penalty0
  1247--1250, 2014.

\bibitem[Chauhan et~al.(2018)Chauhan, Ghanshala, and
  Joshi]{chauhan2018convolutional}
Rahul Chauhan, Kamal~Kumar Ghanshala, and R.~C. Joshi.
\newblock Convolutional neural network ({CNN}) for image detection and
  recognition.
\newblock In \emph{2018 First International Conference on Secure Cyber
  Computing and Communication ({ICSCCC})}, pages 278--282. IEEE, 2018.

\bibitem[Ciresan et~al.(2011)Ciresan, Meier, Masci, Gambardella, and
  Schmidhuber]{ciresan2011flexible}
Dan~Claudiu Ciresan, Ueli Meier, Jonathan Masci, Luca~Maria Gambardella, and
  J{\"u}rgen Schmidhuber.
\newblock Flexible, high performance convolutional neural networks for image
  classification.
\newblock In \emph{Twenty-Second International Joint Conference on Artificial
  Intelligence}, 2011.

\bibitem[Collinson(1998)]{collinson1998bombers}
P.~Collinson.
\newblock Of bombers, radiologists, and cardiologists: time to {ROC}.
\newblock \emph{Heart}, 80\penalty0 (3):\penalty0 215--217, 1998.

\bibitem[Das et~al.(2018)Das, Catthoor, and Schaafsma]{das2018heartbeat}
Anup Das, Francky Catthoor, and Siebren Schaafsma.
\newblock Heartbeat classification in wearables using multi-layer perceptron
  and time-frequency joint distribution of {ECG}.
\newblock In \emph{Proceedings of the 2018 IEEE/ACM International Conference on
  Connected Health: Applications, Systems and Engineering Technologies}, pages
  69--74, 2018.

\bibitem[Demirkaya et~al.(2020)Demirkaya, Chen, and
  Oymak]{demirkaya2020exploring}
Ahmet Demirkaya, Jiasi Chen, and Samet Oymak.
\newblock Exploring the role of loss functions in multiclass classification.
\newblock In \emph{2020 54th Annual Conference on Information Sciences and
  Systems (CISS)}, pages 1--5. IEEE, 2020.

\bibitem[Desai and Shah(2021)]{desai2021anatomization}
Meha Desai and Manan Shah.
\newblock An anatomization on breast cancer detection and diagnosis employing
  multi-layer perceptron neural network ({MLP}) and convolutional neural
  network ({CNN}).
\newblock \emph{Clinical eHealth}, 4:\penalty0 1--11, 2021.

\bibitem[Diakogiannis et~al.(2020)Diakogiannis, Waldner, Caccetta, and
  Wu]{diakogiannis2020resunet}
Foivos~I. Diakogiannis, Fran{\c{c}}ois Waldner, Peter Caccetta, and Chen Wu.
\newblock {ResUNet-a}: A deep learning framework for semantic segmentation of
  remotely sensed data.
\newblock \emph{{ISPRS} Journal of Photogrammetry and Remote Sensing},
  162:\penalty0 94--114, 2020.

\bibitem[Dua and Graff(2017)]{Dua:2019}
Dheeru Dua and Casey Graff.
\newblock {UCI} machine learning repository, 2017.
\newblock URL \url{http://archive.ics.uci.edu/ml}.

\bibitem[Feng et~al.(2018)Feng, Kittler, Awais, Huber, and Wu]{feng2018wing}
Zhen-Hua Feng, Josef Kittler, Muhammad Awais, Patrik Huber, and Xiao-Jun Wu.
\newblock Wing loss for robust facial landmark localisation with convolutional
  neural networks.
\newblock In \emph{Proceedings of the {IEEE} Conference on Computer Vision and
  Pattern Recognition}, pages 2235--2245, 2018.

\bibitem[Ghosh et~al.(2017)Ghosh, Kumar, and Sastry]{ghosh2017robust}
Aritra Ghosh, Himanshu Kumar, and PS~Sastry.
\newblock Robust loss functions under label noise for deep neural networks.
\newblock In \emph{Proceedings of the AAAI Conference on Artificial
  Intelligence}, volume~31, 2017.

\bibitem[Hellinger(1909)]{hellinger1909neue}
Ernst Hellinger.
\newblock Neue begr{\"u}ndung der theorie quadratischer formen von
  unendlichvielen ver{\"a}nderlichen.
\newblock \emph{Journal f{\"u}r die reine und angewandte Mathematik},
  1909\penalty0 (136):\penalty0 210--271, 1909.

\bibitem[Janocha and Czarnecki(2017)]{janocha2017loss}
Katarzyna Janocha and Wojciech~Marian Czarnecki.
\newblock On loss functions for deep neural networks in classification.
\newblock \emph{arXiv preprint arXiv:1702.05659}, 2017.

\bibitem[Kingma and Ba(2014)]{kingma2014adam}
Diederik~P. Kingma and Jimmy Ba.
\newblock {Adam}: A method for stochastic optimization.
\newblock \emph{arXiv preprint arXiv:1412.6980}, 2014.

\bibitem[Krizhevsky and Hinton(2009)]{krizhevsky2009learning}
Alex Krizhevsky and Geoffrey Hinton.
\newblock Learning multiple layers of features from tiny images.
\newblock 2009.

\bibitem[Kullback and Leibler(1951)]{kullback1951information}
Solomon Kullback and Richard~A. Leibler.
\newblock On information and sufficiency.
\newblock \emph{The Annals of Mathematical Statistics}, 22\penalty0
  (1):\penalty0 79--86, 1951.

\bibitem[LeCun et~al.(1989)LeCun, Boser, Denker, Henderson, Howard, Hubbard,
  and Jackel]{lecun1989backpropagation}
Yann LeCun, Bernhard Boser, John~S. Denker, Donnie Henderson, Richard~E.
  Howard, Wayne Hubbard, and Lawrence~D. Jackel.
\newblock Backpropagation applied to handwritten zip code recognition.
\newblock \emph{Neural Computation}, 1\penalty0 (4):\penalty0 541--551, 1989.

\bibitem[Lin et~al.(2017)Lin, Goyal, Girshick, He, and
  Doll{\'a}r]{lin2017focal}
Tsung-Yi Lin, Priya Goyal, Ross Girshick, Kaiming He, and Piotr Doll{\'a}r.
\newblock Focal loss for dense object detection.
\newblock In \emph{Proceedings of the {IEEE} International Conference on
  Computer Vision}, pages 2980--2988, 2017.

\bibitem[Matthews(1975)]{matthews1975comparison}
Brian~W. Matthews.
\newblock Comparison of the predicted and observed secondary structure of t4
  phage lysozyme.
\newblock \emph{Biochimica et Biophysica Acta (BBA)-Protein Structure},
  405\penalty0 (2):\penalty0 442--451, 1975.

\bibitem[Murphy(2012)]{murphy2012machine}
Kevin~P. Murphy.
\newblock \emph{Machine Learning: A Probabilistic Perspective}.
\newblock MIT Press, 2012.

\bibitem[Muthukumar et~al.(2021)Muthukumar, Narang, Subramanian, Belkin, Hsu,
  and Sahai]{Muthukumar2021ClassificationVR}
Vidya Muthukumar, Adhyyan Narang, Vignesh Subramanian, Mikhail Belkin,
  Daniel~J. Hsu, and Anant Sahai.
\newblock Classification vs regression in overparameterized regimes: Does the
  loss function matter?
\newblock \emph{Journal of Machine Learning Research}, 22:\penalty0
  222:1--222:69, 2021.

\bibitem[Potghan et~al.(2018)Potghan, Rajamenakshi, and
  Bhise]{potghan2018multi}
Sneha Potghan, R.~Rajamenakshi, and Archana Bhise.
\newblock Multi-layer perceptron based lung tumor classification.
\newblock In \emph{2018 Second International Conference on Electronics,
  Communication and Aerospace Technology (ICECA)}, pages 499--502. IEEE, 2018.

\bibitem[Qi et~al.(2020)Qi, Du, Siniscalchi, Ma, and Lee]{qi2020mean}
Jun Qi, Jun Du, Sabato~Marco Siniscalchi, Xiaoli Ma, and Chin-Hui Lee.
\newblock On mean absolute error for deep neural network based vector-to-vector
  regression.
\newblock \emph{IEEE Signal Processing Letters}, 27:\penalty0 1485--1489, 2020.

\bibitem[Rumelhart et~al.(1985)Rumelhart, Hinton, and
  Williams]{rumelhart1985learning}
David~E. Rumelhart, Geoffrey~E. Hinton, and Ronald~J. Williams.
\newblock Learning internal representations by error propagation.
\newblock Technical report, University of California San Diego, La Jolla
  Institute for Cognitive Science, 1985.

\bibitem[Tabik et~al.(2017)Tabik, Peralta, Herrera-Poyatos, and
  Herrera~Triguero]{tabik2017snapshot}
Siham Tabik, Daniel Peralta, Andres Herrera-Poyatos, and Francisco
  Herrera~Triguero.
\newblock A snapshot of image pre-processing for convolutional neural networks:
  case study of {MNIST}.
\newblock 2017.

\bibitem[Vashisth et~al.(2020)Vashisth, Dhall, and
  Saraswat]{vashisth2020chronic}
Shubham Vashisth, Ishika Dhall, and Shipra Saraswat.
\newblock Chronic kidney disease ({CKD}) diagnosis using multi-layer perceptron
  classifier.
\newblock In \emph{2020 10th International Conference on Cloud Computing, Data
  Science \& Engineering (Confluence)}, pages 346--350. IEEE, 2020.

\bibitem[Vaswani et~al.(2017)Vaswani, Shazeer, Parmar, Uszkoreit, Jones, Gomez,
  Kaiser, and Polosukhin]{vaswani2017attention}
Ashish Vaswani, Noam Shazeer, Niki Parmar, Jakob Uszkoreit, Llion Jones,
  Aidan~N. Gomez, {\L}ukasz Kaiser, and Illia Polosukhin.
\newblock Attention is all you need.
\newblock \emph{Advances in Neural Information Processing Systems}, 30, 2017.

\bibitem[Wan et~al.(2018)Wan, Liang, Zhang, and Guizani]{wan2018deep}
Shaohua Wan, Yan Liang, Yin Zhang, and Mohsen Guizani.
\newblock Deep multi-layer perceptron classifier for behavior analysis to
  estimate parkinson's disease severity using smartphones.
\newblock \emph{IEEE Access}, 6:\penalty0 36825--36833, 2018.

\bibitem[Wang et~al.(2019)Wang, Hua, Kodirov, and Robertson]{wang2019imae}
Xinshao Wang, Yang Hua, Elyor Kodirov, and Neil~M. Robertson.
\newblock {IMAE} for noise-robust learning: Mean absolute error does not treat
  examples equally and gradient magnitude's variance matters.
\newblock \emph{arXiv preprint arXiv:1903.12141}, 2019.

\bibitem[Wang et~al.(2016)Wang, Chang, Yang, Liu, and Huang]{wang2016studying}
Zhangyang Wang, Shiyu Chang, Yingzhen Yang, Ding Liu, and Thomas~S. Huang.
\newblock Studying very low resolution recognition using deep networks.
\newblock In \emph{Proceedings of the {IEEE} Conference on Computer Vision and
  Pattern Recognition}, pages 4792--4800, 2016.

\bibitem[Xiao et~al.(2017)Xiao, Rasul, and Vollgraf]{xiao2017fashion}
Han Xiao, Kashif Rasul, and Roland Vollgraf.
\newblock Fashion-{MNIST}: a novel image dataset for benchmarking machine
  learning algorithms.
\newblock \emph{arXiv preprint arXiv:1708.07747}, 2017.

\bibitem[Yessou et~al.(2020)Yessou, Sumbul, and Demir]{yessou2020comparative}
Hichame Yessou, Gencer Sumbul, and Beg{\"u}m Demir.
\newblock A comparative study of deep learning loss functions for multi-label
  remote sensing image classification.
\newblock In \emph{IGARSS 2020-2020 IEEE International Geoscience and Remote
  Sensing Symposium}, pages 1349--1352. IEEE, 2020.

\bibitem[Yule(1912)]{yule1912methods}
G.~Udny Yule.
\newblock On the methods of measuring association between two attributes.
\newblock \emph{Journal of the Royal Statistical Society}, 75\penalty0
  (6):\penalty0 579--652, 1912.

\bibitem[Yulita et~al.(2018)Yulita, Rosadi, Purwani, and
  Suryani]{yulita2018multi}
Intan~Nurma Yulita, Rudi Rosadi, Sri Purwani, and Mira Suryani.
\newblock Multi-layer perceptron for sleep stage classification.
\newblock In \emph{Journal of Physics: Conference Series}, volume 1028, page
  012212. IOP Publishing, 2018.

\end{thebibliography}
\newpage
\clearpage
\appendix

\section{Architectures}\label{appendix:architectures}
\begin{figure}[!h]
	\centering
  \scalebox{.85}{\subfigure[CNN architecture for {\Mnist}. Dropout is applied after the \nth{1} and \nth{2} conv layer (\(p = 0.2\)), as well as after the dense layer (\(p = 0.2\)). The \nth{1} conv layer has \(5 \times 5\) filters.]{\label{fig:mnist-cnn}\includegraphics[width=.65\linewidth]{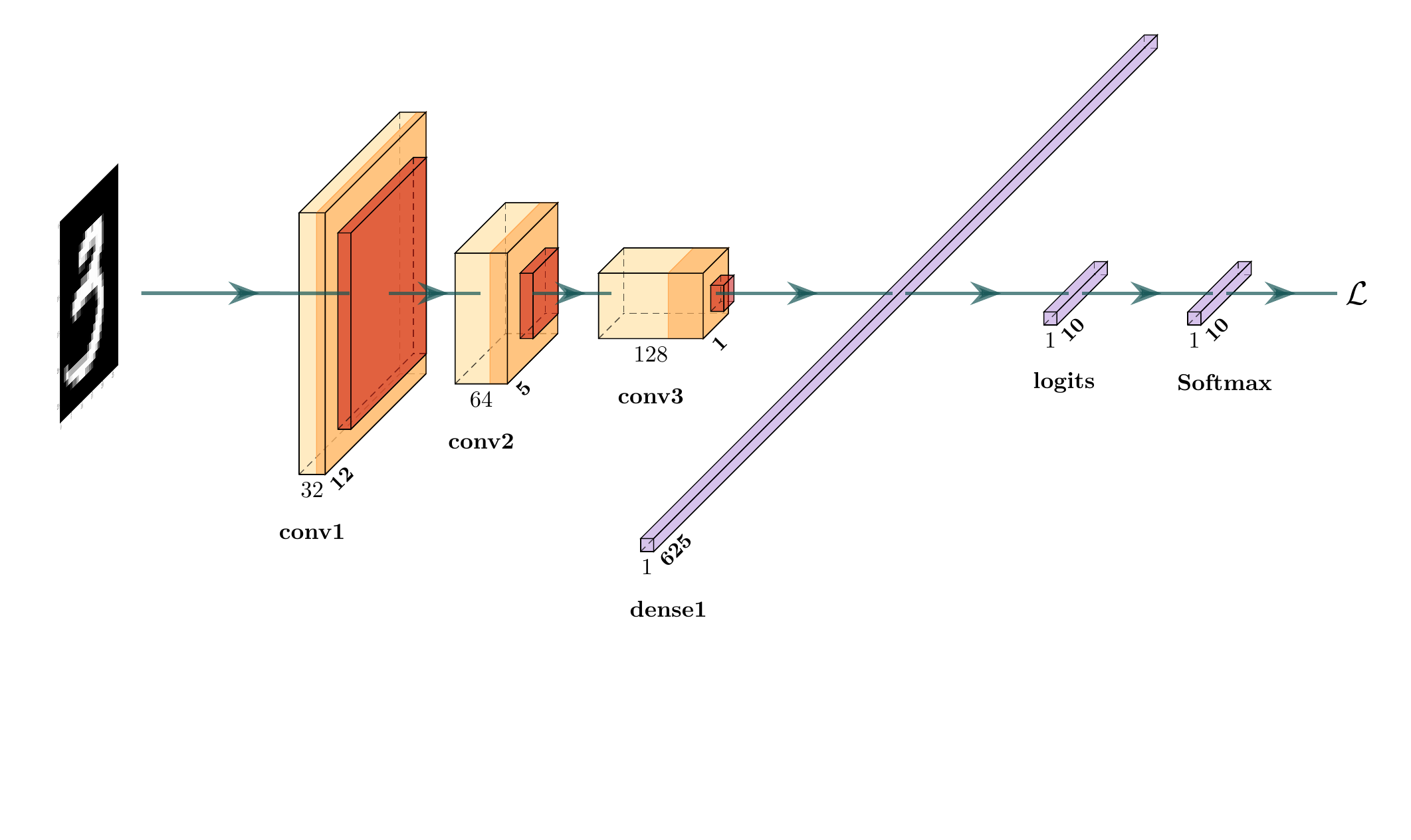}}}
  \scalebox{.85}{\subfigure[CNN architecture for {\Cifar}. Dropout is applied after both Max-Pool layers (\(p = 0.25\)), after the \nth{4} conv layer (\(p = 0.25\)) and after the dense layer (\(p = 0.5\)). The \nth{1} and \nth{2} conv layer have \(5 \times 5\) filters.]{\label{fig:cifar-cnn}\includegraphics[width=.65\linewidth]{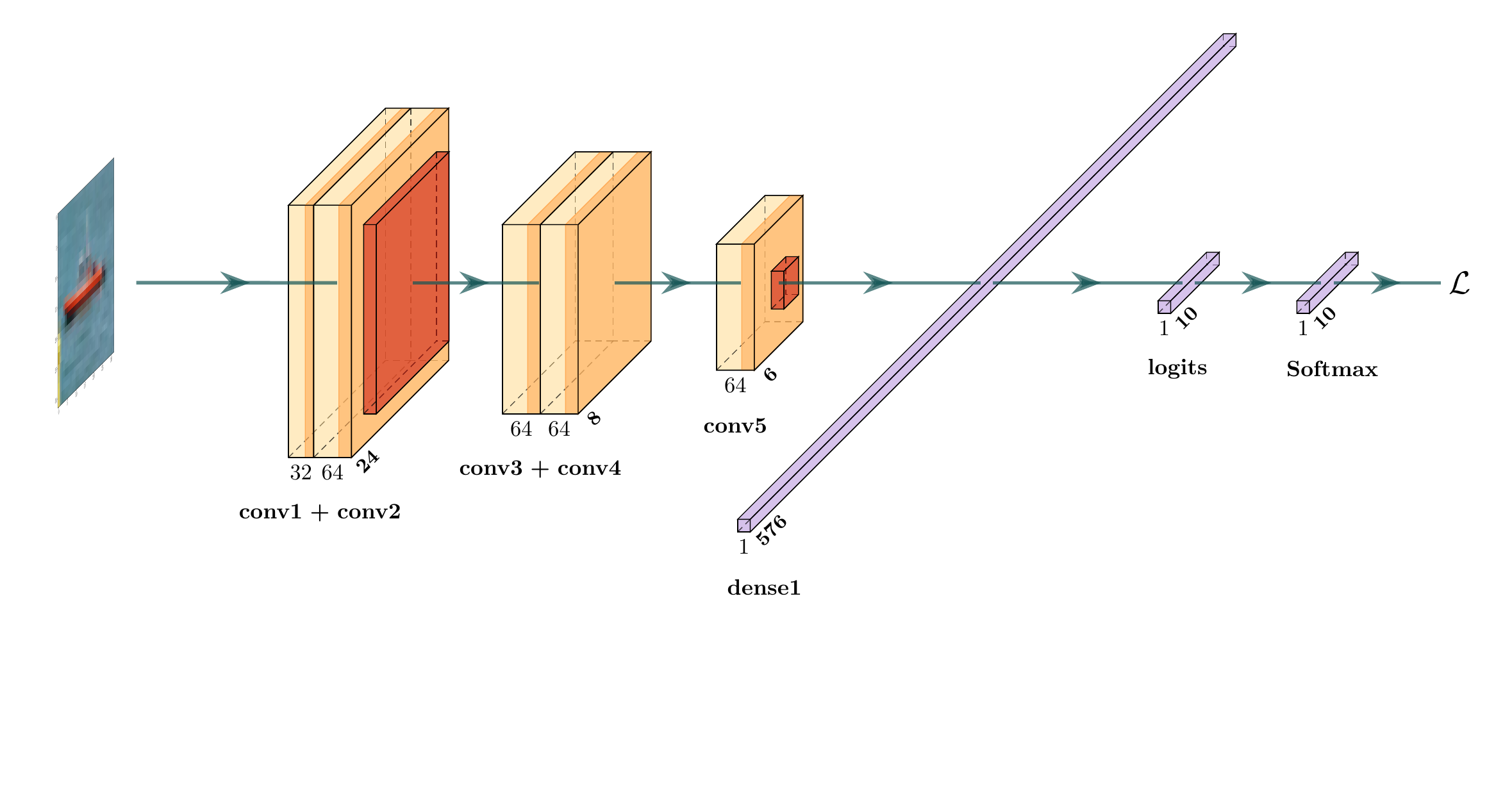}}}
  \scalebox{.85}{\subfigure[CNN architecture for {\CharFontIm}. Dropout is applied after the \nth{2}, \nth{3} and \nth{4} conv layer (\(p = 0.25\)), as well as after the \nth{1} and \nth{2} dense layers (\(p = 0.5\)). Filters have size \(5 \times 5\) for the \nth{1} and \nth{2} conv layers.]{\label{fig:charfontim-cnn}\includegraphics[width=.65\linewidth]{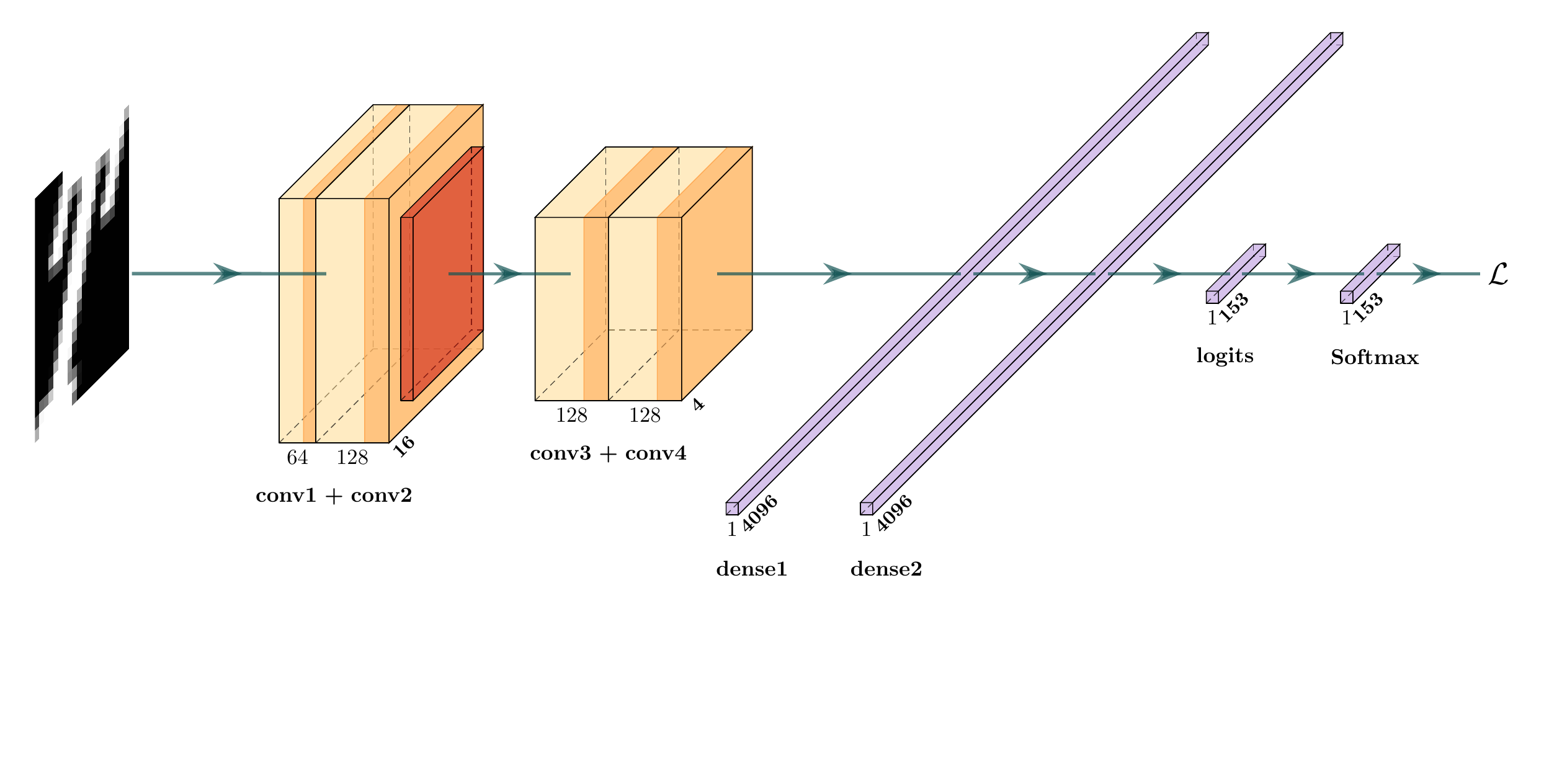}}}
	\caption{CNN architectures. Red layers are Max-Pool layers of size \(2 \times 2\). If not specified, the filter size of a conv layer is \(3 \times 3\).}
	\label{fig:mnist-cifar-charfontim-cnn}
\end{figure}
\begin{figure}[!h]
	\centering
	\newcommand{\inputnum}{3}
	\newcommand{\hiddennum}{5}
	\newcommand{\outputnum}{3}
	\begin{tikzpicture}
	\tikzset{
    position label/.style={
       below = 3pt,
       text height = 1.5ex,
       text depth = 1ex
    },
   brace/.style={
     decoration={brace, mirror},
     decorate
   }
}

	\node[circle,
		minimum size=8mm,
		line width=1pt,
		draw=black,
		fill=orange!30
	] (Input-1) at (0,-1) {};

	\node[
	] (Input-2) at (0,-2) {\Large\(\vdots\)};
	
	\node[circle,
		minimum size=8mm,
		line width=1pt,
		draw=black,
		fill=orange!30
	] (Input-3) at (0,-3) {};



	\node[circle,
		minimum size = 8mm,
		line width=1pt,
		draw=black,
		fill=teal!50,
		yshift=(\hiddennum-\inputnum)*5 mm
	] (Hidden1-1) at (2.5,-1) {};
	\node[circle,
		minimum size = 8mm,
		line width=1pt,
		draw=black,
		fill=teal!50,
		yshift=(\hiddennum-\inputnum)*5 mm
	] (Hidden1-2) at (2.5,-2) {};
	\node[
		yshift=(\hiddennum-\inputnum)*5 mm
	] (Hidden1-3) at (2.5,-3) {\Large\(\vdots\)};
	\node[circle,
		minimum size = 8mm,
		line width=1pt,
		draw=black,
		fill=teal!50,
		yshift=(\hiddennum-\inputnum)*5 mm
	] (Hidden1-4) at (2.5,-4) {};
	\node[circle,
		minimum size = 8mm,
		line width=1pt,
		draw=black,
		fill=teal!50,
		yshift=(\hiddennum-\inputnum)*5 mm
	] (Hidden1-5) at (2.5,-5) {};
	

	\node[circle,
		minimum size = 8mm,
		line width=1pt,
		draw=black,
		fill=teal!50,
		yshift=(\hiddennum-\inputnum)*5 mm
	] (Hidden2-1) at (5,-1) {};
	\node[circle,
		minimum size = 8mm,
		line width=1pt,
		draw=black,
		fill=teal!50,
		yshift=(\hiddennum-\inputnum)*5 mm
	] (Hidden2-2) at (5,-2) {};
	\node[
		yshift=(\hiddennum-\inputnum)*5 mm
	] (Hidden2-3) at (5,-3) {\Large\(\vdots\)};
	\node[circle,
		minimum size = 8mm,
		line width=1pt,
		draw=black,
		fill=teal!50,
		yshift=(\hiddennum-\inputnum)*5 mm
	] (Hidden2-4) at (5,-4) {};
	\node[circle,
		minimum size = 8mm,
		line width=1pt,
		draw=black,
		fill=teal!50,
		yshift=(\hiddennum-\inputnum)*5 mm
	] (Hidden2-5) at (5,-5) {};

	\draw[brace] (Hidden1-5.south) -- node[position label, pos=0.5] {4 layers} (Hidden2-5.south);
	\node (output-neurons) at (7.5,-4) {Output layer};


	\node[circle,
		minimum size = 8mm,
		line width=1pt,
		draw=black,
		fill=purple!50,
		yshift=(\outputnum-\inputnum)*5 mm
	] (Output-1) at (7.5,-1) {};
	\node[
		yshift=(\outputnum-\inputnum)*5 mm
	] (Output-2) at (7.5,-2) {\Large\(\vdots\)};
	\node[circle,
		minimum size = 8mm,
		line width=1pt,
		draw=black,
		fill=purple!50,
		yshift=(\outputnum-\inputnum)*5 mm
	] (Output-3) at (7.5,-3) {};
	

	\node[circle,
		minimum size = 8mm,
		line width=1pt,
		draw=black,
		fill=xkcdNiceBlue!50,
		yshift=(\outputnum-\inputnum)*5 mm
	] (Softmax-1) at (10,-1) {};
	\node[
		yshift=(\outputnum-\inputnum)*5 mm
	] (Softmax-2) at (10,-2) {\Large\(\vdots\)};
	\node[circle,
		minimum size = 8mm,
		line width=1pt,
		draw=black,
		fill=xkcdNiceBlue!50,
		yshift=(\outputnum-\inputnum)*5 mm
	] (Softmax-3) at (10,-3) {};

	\foreach \i in {1,...,\inputnum}
	{
		\foreach \j in {1,...,\hiddennum}
		{
			\draw[->, shorten >=1pt] (Input-\i) -- (Hidden1-\j);
		}
	}
	
	\foreach \i in {1,...,\hiddennum}
	{
		\foreach \j in {1,...,\hiddennum}
		{
			\draw[->, shorten >=1pt] (Hidden1-\i) -- (Hidden2-\j);
		}
	}
	
	\path (Hidden1-1) -- node[midway,above=1.5em] {\hspace{0.2in}\footnotesize ReLU activations \(\cdots\)} (Hidden2-1);

	\foreach \i in {1,...,\hiddennum}
	{
		\foreach \j in {1,...,\outputnum}
		{
			\draw[->, shorten >=1pt] (Hidden2-\i) -- (Output-\j);
		}
	}
	
	\foreach \i in {1,...,\outputnum}
	{
		\draw[->, shorten >=1pt] (Output-\i) -- (Softmax-\i);
	}
	
	\path (Output-1) -- node[midway,above=1.5em] {\scriptsize Softmax} (Softmax-1);
	
	\path (Output-1) -- node[midway,below=-0.5em] {\(\vdots\)} (Softmax-1);
	\path (Output-2) -- node[midway,above=0.1em]  {\(\bm{o}\)} (Softmax-2);
	\path (Output-2) -- node[midway,below=-0.1em] {\(\vdots\)} (Softmax-2);

	\draw[<-, shorten <=1pt] (Input-1) -- ++(-1,0) node[left]{\(x_{i}^{(1)}\)};
	\draw[<-, shorten <=1pt] (Input-2) -- ++(-1,0) node[left]{\Large\(\vdots\)};
	\draw[<-, shorten <=1pt] (Input-3) -- ++(-1,0) node[left]{\(x_{i}^{(d)}\)};
	
	\draw[->, shorten <=1pt] (Softmax-1) -- ++(1,0) node[right]{\(\hat{y}_{i}^{(1)}\)};
	\draw[->, shorten <=1pt] (Softmax-2) -- ++(1,0) node[right]{\Large\(\vdots\)};
	\draw[->, shorten <=1pt] (Softmax-3) -- ++(1,0) node[right]{\(\hat{y}_{i}^{(c)}\)};
	\end{tikzpicture}
	\caption{Our dense architecture. Hidden layers have a teal color. The first dense layer has 32 neurons, the second and third 64, and the fourth 32. This is followed by a fifth dense layer (red-ish colored) having 2 output neurons. Dropout is applied after the first 3 hidden layers (\(p = 0.4 \rightarrow 0.2 \rightarrow 0.2\)). Note that this representation is not confined to a fixed number of classes or neurons; the vertical dots indicate that it is indeed conceptual.}
	\label{fig:dense-architecture}
\end{figure}
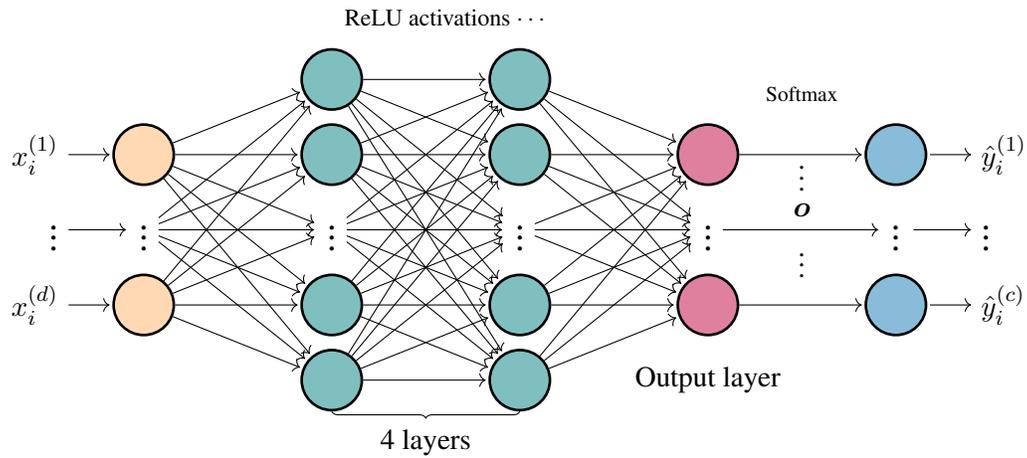

\newpage

\section{Accuracy Progressions}\label{appendix:accuracy-progressions}
\begin{figure}[!h]
	\centering
  \subfigure{\label{fig:mush-accuracies}\includegraphics[width=.35\textwidth]{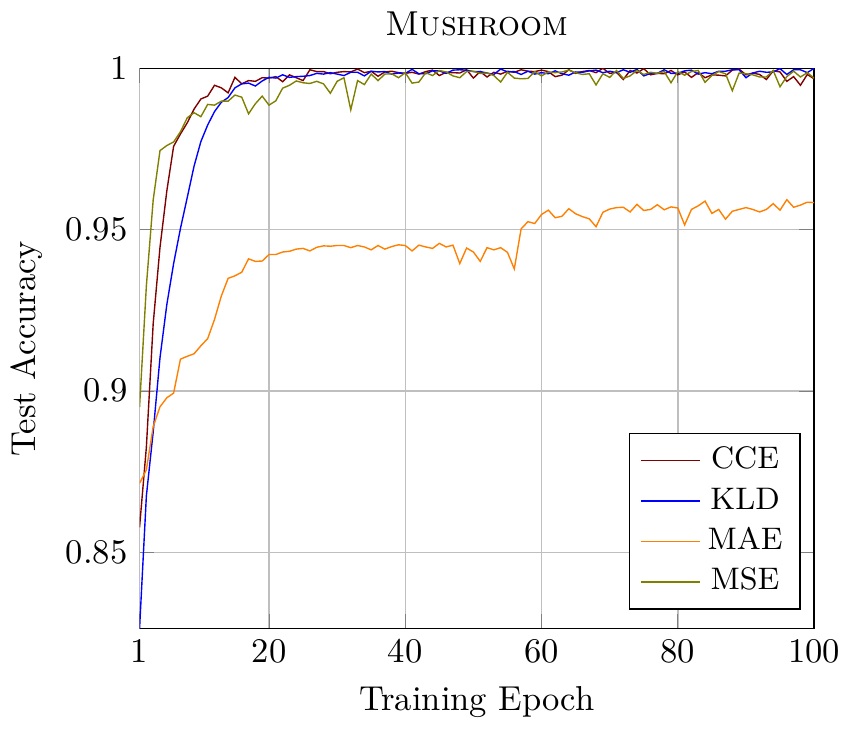}}
  \subfigure{\label{fig:phish-accuracies}\includegraphics[width=.35\textwidth]{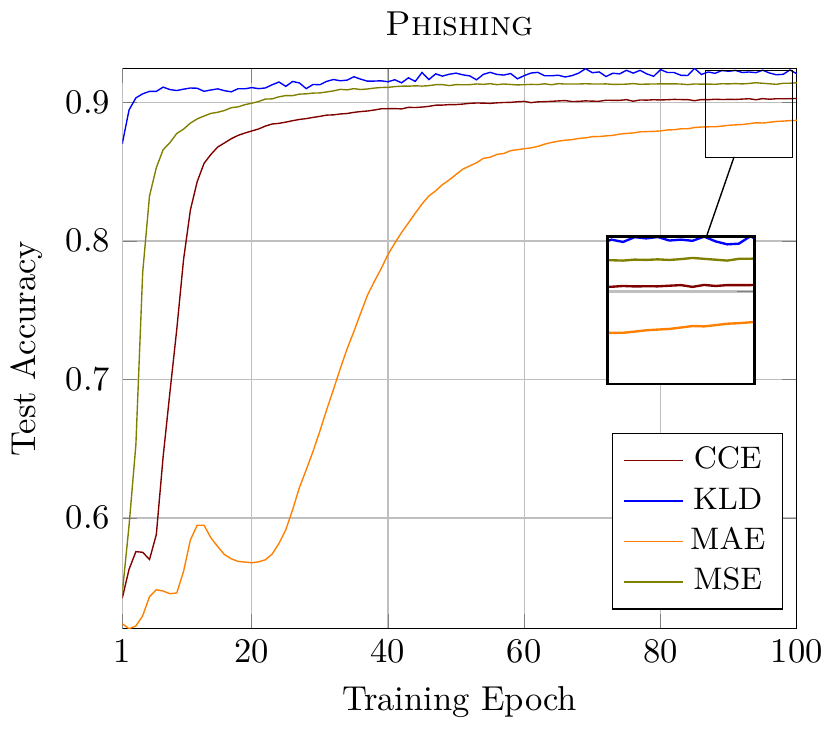}}
  \subfigure{\label{fig:mnist-accuracies}\includegraphics[width=.35\textwidth]{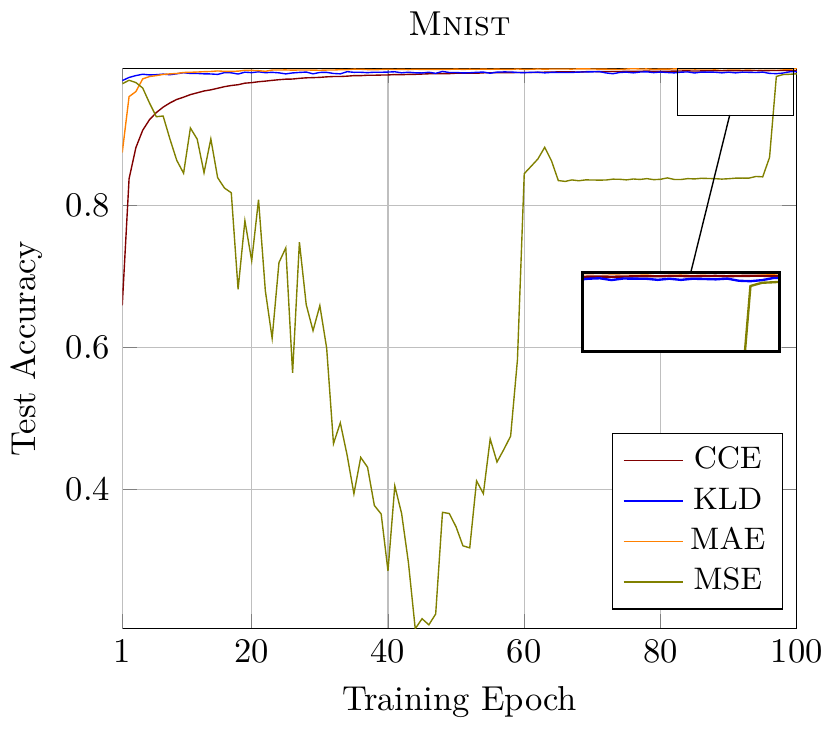}}
  \subfigure{\label{fig:fashion-accuracies}\includegraphics[width=.35\textwidth]{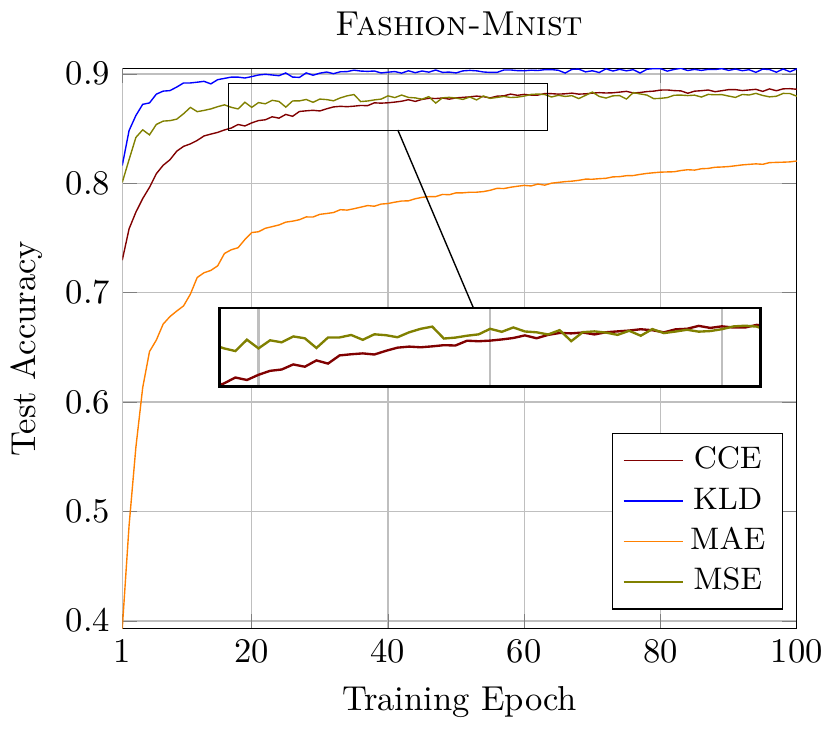}}
  \subfigure{\label{fig:cifar-accuracies}\includegraphics[width=.35\textwidth]{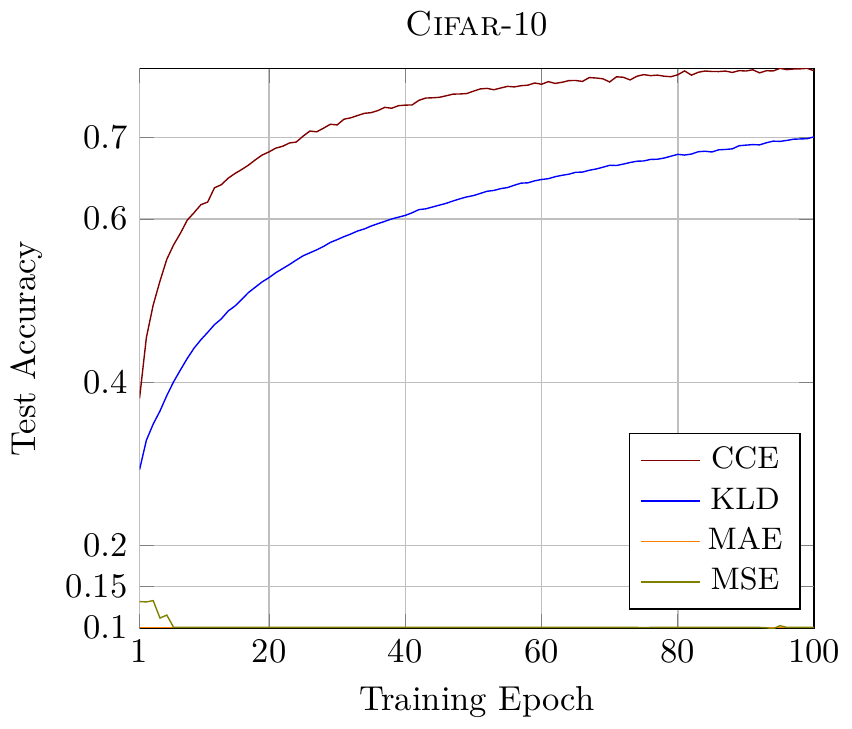}}
  \subfigure{\label{fig:charfontim-accuracies}\includegraphics[width=.35\textwidth]{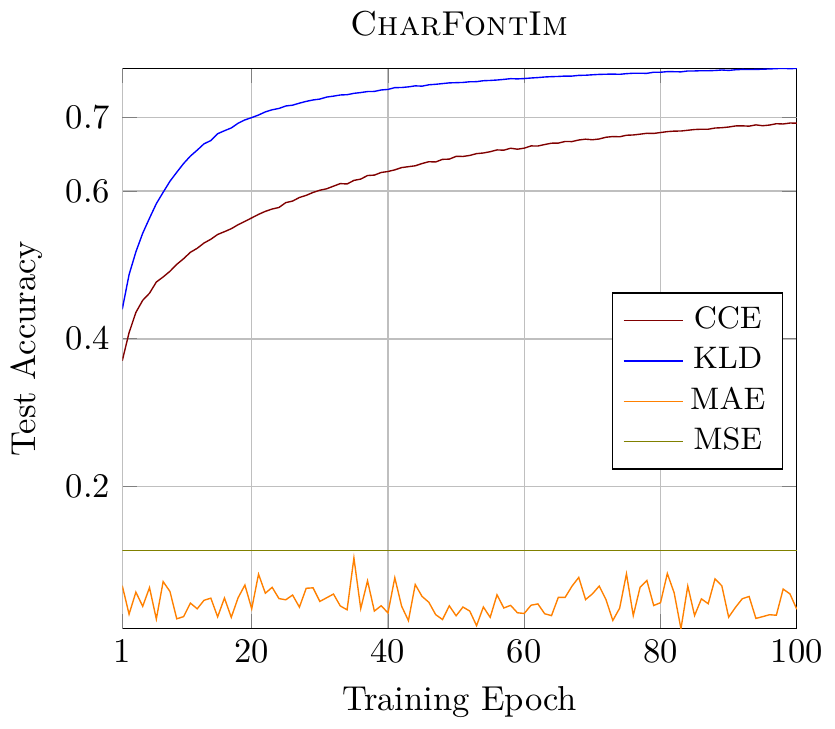}}
	\caption{Test accuracies on each data set (across 6 iterations). Points of interest are inset.}
	\label{fig:accuracy-progressions}
\end{figure}

\end{document}